\documentclass[10pt,twocolumn,letterpaper]{article}

\usepackage[pagenumbers]{cvpr} 

\usepackage[accsupp]{axessibility}
\usepackage{graphicx}
\usepackage{amsmath}
\usepackage{amssymb}
\usepackage{booktabs}
\usepackage{wasysym}
\usepackage{siunitx}  

\usepackage{array}
\usepackage{multirow}

\usepackage[pagebackref,breaklinks,colorlinks]{hyperref}

\DeclareUnicodeCharacter{0013}{ }

\usepackage[capitalize]{cleveref}
\crefname{section}{Sec.}{Secs.}
\Crefname{section}{Section}{Sections}
\Crefname{table}{Table}{Tables}
\crefname{table}{Tab.}{Tabs.}

\def\cvprPaperID{27} 
\def\confName{CVPR}
\def\confYear{2022}

\begin{document}

\title{Perfusion assessment via local remote photoplethysmography (rPPG)}

\author{\smallskip{Benjamin Kossack$^{1}$ \qquad Eric Wisotzky$^{1,2}$  \qquad Peter Eisert$^{1,2}$ \qquad Sebastian P. Schraven$^{3}$} \\ \smallskip{Brigitta Globke$^{4}$ \qquad Anna Hilsmann$^{1}$}\\
$^{1}$Fraunhofer Heinrich Hertz Institute, Berlin \qquad $^{2}$Humboldt-Universität zu Berlin, Germany \\
$^{3}$Department of Otorhinolaryngology, Head and Neck Surgery “Otto Körner”,\\ Rostock University Medical Center, Rostock, Germany \\
$^{4}$Department of Surgery, Charité-Universitätsmedizin Berlin,
Berlin, Germany\\
{\tt\small benjamin.kossack@hhi.fraunhofer.de}
}
\maketitle

\begin{abstract}
    This paper presents an approach to assess the perfusion of visible human tissue from RGB video files. We propose metrics derived from remote photoplethysmography (rPPG) signals to detect whether a tissue is adequately supplied with blood. The perfusion analysis is done in three different scales, offering a flexible approach for different applications. We perform a plane-orthogonal-to-skin rPPG independently for locally defined regions of interest on each scale. From the extracted signals, we derive the signal-to-noise ratio, magnitude in the frequency domain, heart rate, perfusion index as well as correlation between specific rPPG signals in order to locally assess the perfusion of a specific region of human tissue. We show that locally resolved rPPG has a broad range of applications. As exemplary applications, we present results in intraoperative perfusion analysis and visualization during skin and organ transplantation as well as an application for liveliness assessment for the detection of presentation attacks to authentication systems.
\end{abstract}

\section{Introduction}
\label{sec:intro}
Perfusion assessment and vital sign monitoring have been addressed in recent years with a vast amount of possible solutions. The range of proposed applications last from clinical diagnostics and monitoring up to liveliness analysis in security applications.

Especially, the assessment of perfusion is challenging as an analysis of the blood flow requires temporally continuous, and at the same time, locally resolved information. Most solutions do not meet all these requirements. Global vital signs can be measured objectively and continuously using a monitoring system with sensors attached to the patient. Heart rate (HR) measurement is usually based on photoplethysmography (PPG) \cite{Zaunseder2018} measuring optically the light transmission and reflectance caused by blood flow \cite{Tamura2014} with a sensor directly placed on the skin of the patient. The same principle is used in remote photoplethysmography (rPPG), which allows contactless measurements of the HR with a regular RGB camera\cite{Poh2010,Wang2017,Kossack2019,Yang2021}. The majority of rPPG-related literature addresses the extraction of global vital signs. Only little work has been published on locally resolved rPPG signals to, \eg, analyze the blood flow through human tissue, mainly in the human face \cite{YangJunGuthierBenjamin2015,Zaunseder2018,Kossack2019,KossackPad2019}.

Although the assessment of optimal blood flow is essential for many medical procedures, \eg~tissue or organ transplantation, in order to evaluate the surgical success, the current gold standard is still subjective clinical assessment of the situs and patient's condition at the bedside. Existing monitoring techniques have been reported in the past, but due to several disadvantages, \eg, invasiveness and potential complications, they failed to gain acceptance \cite{Swartz1988,King2000,Mothes2004}. 
A promising non-invasive, non-contact and non-ionising but cost-intensive method is hyperspectral imaging \cite{yoon2022hyperspectral,muhle2021comparison,thiem2021hyperspectral,wisotzky2020validation}. 
Further, telemedicine as a remote service for diagnosis and medical monitoring is a fast-growing public health sector. Especially, the COVID-19 pandemic moved it into focus. Unfortunately, telehealth does not allow measuring the vital signs using patient contact. However, the face of the patient can be an essential source of information about vital signs and well-being. Therefore, it would be beneficial to directly determine the patient's condition from the face in the video stream. Thus, there is a growing need for an objective, reliable, and examiner-independent method to safely assess perfusion and vital signs in many medical applications, ranging from intraoperative monitoring as well as telehealth applications.

In addition to clinical assessment, rPPG can also be used in security applications for liveliness analysis in order to detect presentation attacks (PAD)~\cite{Suh2016,Heusch2019,Li2017,Nowara2017,KossackPad2019}. The use of a facial recognition system for authentication has become widespread. Biometric authentication systems based on facial recognition are already used in border security systems and unlock smartphones. Although widely used and highly accurate, facial recognition algorithms suffer from vulnerability to simple spoofing attacks by impersonating a target victim holding the desired authorization~\cite{Marasco2016} through the use of, \eg, photos or masks~\cite{Galbally2014}. RPPG methods can be used to detect if the whole face is covered by a mask or not\cite{Li2017}. In addition, locally resolved rPPG analysis, as proposed in this paper, is able to detect partial face masks.

In this paper, we propose a method for locally (i.e.~spatially) resolved rPPG analysis for the assessment of the blood flow through human tissue from videos with the aim to visualize tissue perfusion quality as well as blood flow. In our approach, the perfusion can be assessed at different scales depending on the chosen application: globally for an overall perfusion assessment of, \eg, implanted tissue, region-based to detect poorly or non-perfused regions, and locally to visualize the blood flow. We introduce metrics derived from rPPG to detect whether a tissue is adequately supplied with blood. We compute these multiple parameters from the local rPPG signal, which are then utilized to assess the perfusion. Here, we define three parameters to analyze the perfusion in a region of interest (ROI) based on the rPPG signal.

The remainder of this paper is organized as follows. In \cref{sec:relwork}, we review existing related work on remote photoplethysmography for vital sign and perfusion analysis. The proposed method and the dataset are described in \cref{sec:method} and \cref{sec:data}, respectively. In \cref{sec:app}, we introduce different applications for each presented scale. Then, in \cref{sec:discussion} the results are discussed, while \cref{sec:conclusion} concludes the paper.

\section{Related Work}
\label{sec:relwork}
The measurement of the HR\footnote{Heart rate and pulse rate have been used synonymously in the literature, although they are not precisely the same \cite{Kroemer2010,Rapczynski2017,Tulyakov_2016_CVPR}. To be precise, according to medical definitions, the rPPG signal can be used to extract the pulse rate signal, but we will use the term HR in this paper.} is usually based on the optical measuring technique photoplethysmography (PPG) \cite{Zaunseder2018}. The principle is based on human blood circulation and the fact that blood absorbs more light than surrounding tissue. Thus, variations in blood volume affect light transmission or reflectance accordingly\cite{Tamura2014}. A photoplethysmography (PPG) sensor (commonly used to measure the human pulse rate) is placed directly on the skin and optically detects the changes in blood volume\cite{Tamura2014}. The same principle is used in remote photoplethysmography (rPPG), which allows contactless measurements of the HR with a regular RGB camera\cite{Zaunseder2018}: the blood flow through the human circulatory system leads to a continuous change in skin color, which is analyzed by rPPG techniques to determine the HR \cite{Poh2010,DeHaan2013,Wang2017,Tulyakov_2016_CVPR}. 

To robustly extract an rPPG signal regardless of the subject's skin tone and illumination color (non-white illumination), a chrominance-based calculation (CHROM) of the rPPG signal has been developed \cite{DeHaan2013} for pre-processing the input frames. Similarly, the Plane-Orthogonal-to-Skin Transformation (POS)\cite{Wang2017} projects a three-channel (R-G-B) image onto a plane orthogonal to the [1,~1,~1] direction to create a two-channel image. These two channels are then fused to the desired rPPG signal.

As global model-based methods can be affected, \eg, by noise, compression artifacts, or masking, several of the latest rPPG related publications use deep neural networks to extract the HR from video \cite{Chen2018,Yu_2019_ICCV,Yu}. Yang \etal~\cite{Yang2021} compared three neural networks (Deepphys~\cite{Chen2018}, rPPGNet~\cite{Yu_2019_ICCV}, and Physnet~\cite{Yu}) with model-based approaches (independent component analysis (ICA)~\cite{Poh2010}, CHROM~\cite{DeHaan2013}, and POS~\cite{Wang2017}) on the publicly available UBFC-rPPG dataset \cite{bobbia2019unsupervised}. In these experiments, the deep-learning-based approaches outperform the model-based approaches under constant lighting conditions, but model-based approaches (ICA, CHROM, and POS) show more accurate and robust results than neuronal networks in varying lighting conditions \cite{Yang2021}. The recent ICCV Vision-for-Vitals (V4V) challenge~\cite{Revanur_2021_ICCV} comparing different model-based as well as network-based approaches~\cite{kossack2021, gideon2021estimating,Hill_2021_ICCV,Revanur_2021_ICCV} for the measurement of the HR, shows similar results. 

Recently, rPPG signals have also been analyzed locally and visualized based on amplitude, velocity, or signal-to-noise ratio (SNR) map\cite{YangJunGuthierBenjamin2015,Zaunseder2018}. Especially, the blood flow has been analyzed in videos showing a human face \cite{YangJunGuthierBenjamin2015,Kossack2019, KossackPad2019}. The blood flow velocity is calculated from the relative phase shift of the frequency component corresponding to the HR in the frequency domain. These methods assume that the difference between neighboring phase values directly corresponds to the velocity at this point. Applying a 2D Sobel operator to the calculated phase shifts results to the desired velocity map.
Another method for estimating perfusion speed from videos uses spatially separated rPPG signals; band-pass filtered based on the HR measured with an electrocardiogram (ECG) \cite{Zaunseder2018}. The time delay between specific positions can then be extracted and the pulse wave perfusion speed calculated. However, since the calculation is performed on pixel values, the result is neither a physical speed nor can it be transferred to other data.

Besides medical applications, rPPG analysis has also been applied to detect so-called presentation attacks to authentication systems. In these works, rPPG methods are applied to facial videos in order to detect whether the face is covered by a mask~\cite{Li2017, yu2021transrppg, KossackPad2019}. 

\section{Method}
\label{sec:method}
This chapter presents our approach for rPPG-based perfusion assessment of skin tissue. 
We distinguish between three different scales for the analysis of perfusion and blood flow: (1) \textit{global}, (2) \textit{region-based}, and (3) \textit{local}. The global analysis resembles a one-dimensional rPPG signal extraction from a video recording. In the region-based approach, we subdivide the region of interest (ROI) in the video into sub-regions and analyze these. The local blood flow analysis resolves the rPPG signal on pixel-level in the input RGB video. The three different scales require different prepossessing steps, including annotation of the ROI, image registration, image filtering, and scaling. In the following, we describe our approach for calculating the rPPG signal and the derivation of the perfusion parameters from the rPPG signal. Details on the pre-processing steps for the different scales are explained in sections \cref{sec:global} to \cref{sec:local}.

\subsection{rPPG-based Perfusion Assessment} 
\label{sec:parameter}
In order to extract an rPPG signal for each scale (ROI, region or pixel position), we use the POS transformation \cite{Wang2017} as it showed best results for different lightning conditions \cite{Yang2021}.
For each frame, the red (R), green (G), and blue (B) color values are pixel-wise averaged for each scale and concatenated to a three-dimensional (3D) normalized time signal $\left[r(t), g(t), b(t) \right]$ per scale.
This signal is projected onto a plane orthogonal to an normalized skin tone vector using
\begin{equation}\label{eq:pos1}
S_1(t) = g(t) - b(t)
\end{equation}
and 
\begin{equation}\label{eq:pos2}
S_2(t) = g(t) + b(t) -2r(t),
\end{equation}
The resulting 2D signal is combined into a 1D signal by 
\begin{equation}\label{eq:pos3}
h(t) = S_1(t) + \frac{\sigma(S_1)}{\sigma(S_2)} S_2,
\end{equation}
where $\sigma$ is the standard deviation (SD). This ensures that the resulting signal contains the maximum amount of pulsating component. Details can be found in \cite{Wang2017}. 
The output signal $h(t)$ is normalized and filtered with a fifth order Butterworth digital band-pass filter (between \SI{0.6}{\hertz} and \SI{4.0}{\hertz}), resulting in the rPPG signal $\mathit{rPPG}(t)$. The temporal trend of the rPPG signal is extracted using a sliding window of $t_{win} = \SI{10}{\second}$ with step size $t_{step} = \SI{1}{\second}$.

The HR determination is done in the frequency domain \cite{Kossack2019,Poh2010}. We propose the \textit{sub bin maximum analysis} to determine the heart frequency $f_\mathit{HR}$ of the rPPG signal. For each scale, we transform the time domain rPPG signal into the frequency domain via fast Fourier transform (FFT). Then we identify the maximum sum of the magnitudes of a fundamental frequency and its second harmonic. Thereby, the range from \SI{0.6}{\hertz} to \SI{4.0}{\hertz} (i.e., a HR of \SI{36}{BPM} to \SI{240}{BPM}) is considered as eligible interval of the fundamental frequency. The selected fundamental frequency represents the heart frequency $f_\mathit{HR}$ for the corresponding scale in the analyzed sliding window of size $t_{win}$. This \textit{sub bin maximum analysis} prevents selecting noise-related frequency components as heart frequency. The HR in beats per minute (\SI{}{BPM}) is given by $60 \cdot f_\mathit{HR}$. During $t_{win}$, the subject's heart rate varies, which is referred to as heart rate variability (HRV) \cite{Zaunseder2018}. In the following, the magnitude of $f_\mathit{HR}$ is referred to as $M(f_\mathit{HR})$ (i.e., selected peak on the power spectrum density).

Based on the HR 
and the extracted rPPG signal, we analyze different parameters for perfusion and blood: SNR, magnitude $M$ of the heart frequency, perfusion index (PI), and the reference correlation coefficient $\rho$ between the rPPG signal of ROI and reference region.
The SNR, as well as $M(f_\mathit{HR})$, indicate if the quality of the extracted rPPG signals is sufficient and the measured HR is reliable. The PI immediately reacts to acute and significant changes in blood volume and allows to distinguish between non-perfused and continuously perfused tissue regions quantitatively. The correlation $\rho$ quantifies the relationship between a specific area and a well-perfused reference region. 

\subsubsection{SNR and Magnitude of the rPPG Signal} 
\label{sec:snr}
In order to quantify the rPPG signal quality,  we analyze the SNR, representing the strength of the signal in the frequency domain compared to unwanted noise present in the signal following De Haan \etal~\cite{DeHaan2013}.
\begin{equation}\label{eq:SNR}
\mathit{SNR} = 10 \log_{10} \left( \dfrac{\sum_{k = f_1}^{f_2} \left( U_m(k) M(k) \right)^2} {\sum_{k = f_1}^{f_2} \left( (1 - U_m(k)) M(k) \right)^2} \right), 
\end{equation}
where $M(k)$ is the magnitude of the signal $f(t)$, $f_1$ and $f_2$ define the range in which the SNR is calculated (\eg, 0.6\,Hz to 4.0\,Hz), $k$ is the bin number of the frequency component, and $U_m(k)$ is a binary mask defining the signal interval
\begin{equation}
U_m(k) = 
\left\lbrace
\begin{aligned}
&1,\quad \text{if} \left| f_\mathit{HR} - \Delta f \cdot k\right| \leq (\SI{0.05}{\hertz})\\ 
&1,\quad \text{if} \left| 2f_\mathit{HR} - \Delta f \cdot k\right| \leq (\SI{0.05}{\hertz}),\\
&0,\quad \text{otherwise}
\end{aligned}
\right. 
\end{equation}
where $\Delta f$ is the spectral frequency resolution
\begin{equation}
\Delta f = \dfrac{f_s}{N_\mathit{fft}},
\end{equation}
with the number of FFT points $N_\mathit{fft}$ (i.e.~number of samples) and the sampling frequency $f_s$ of the input signal $f(t)$.
Due to HRV, an interval of \SI{\pm3}{BPM} (\SI{\pm0.05}{\hertz} in frequency domain) around the heart frequency $f_\mathit{HR}$ as well as its second harmonic $2f_\mathit{HR}$ is used to select the signal. The remaining frequency components are classified as noise, i.e.~$(1-U_m(k))$.

Besides the SNR, the magnitude $M(f_\mathit{HR})$ of the heart frequency is relevant for perfusion assessment. Both values are indicators of the strength and quality of the rPPG signal.

\subsubsection{Reference Correlation}
\label{sec:corr}
The reference correlation coefficient $\rho_{ref}$ measures the correlation between the rPPG signals of the ROI $\mathit{rPPG}_{roi}(t)$ and the reference region $\mathit{rPPG}_\mathit{ref}(t)$\cite{kossack2021}. If no reference region can be specified within the image, an externally measured HR can be used to imitate a reference signal by using a sine wave. 

\subsubsection{Perfusion Index}
\label{sec:perfIdx}
Initially, the Perfusion Index (PI) is derived from the PPG signal and represents the ratio of pulsatile on non-pulsatile light absorbance or reflectance of the PPG signal, but it can also be computed from the rPPG signal as a ratio between the DC and AC component\cite{Rasmussen2021}. However, it has to be calculated using only the green color channel values $g(t)$ of the ROI \cite{Rapczynski2017} as the rPPG signal after POS transformation contains no DC component, which would lead to a division by zero.
\begin{equation}\label{eq:pi}
	\mathit{PI} = \dfrac{g_{lp, \mathit{AC}}}{g_{lp, \mathit{DC}}} = \dfrac{\max(g_{lp}(t))}{\mu(g_{lp}(t))},
\end{equation}
where time signal $g(t)$ is low-pass filtered with cutoff frequency $f_{cut}=\SI{20}{Hz}$ leading to $g_{lp}(t)$ and $\mu$ is the mean and max finds the largest value in the time series $g_{lp}(t)$.
The sample rate $f_s$ and the corresponding Nyquist frequency influence $f_{cut}$ and if $f_{cut}=\SI{20}{Hz}$ is not applicable, $0.8 \cdot f_s/2$ has been used. 

\subsection{Global Analysis}
\label{sec:global}
The global analysis allows an overall perfusion assessment of tissue, based on the rPPG signal, the determined HR, and the parameters presented in \cref{sec:parameter}. In the pre-processing steps, an ROI and, if possible, a reference region has to be defined. Based on these areas, the captured frames are motion-compensated 
as breathing, heartbeat, and external physical contact cause slight movements in the video. We determine the rPPG signals $\mathit{rPPG}_{roi}(t)$ and $\mathit{rPPG}_\mathit{ref}(t)$ for the ROI as well as the reference region in order to compute the HR and the parameters for the perfusion assessment (see \cref{sec:parameter}). When skin tissue is the focus of the analyzed video recording, skin segmentation is applied to each frame removing occlusions within the defined ROI (\eg, eyeglasses in facial images). This skin segmentation increases the SNR of the rPPG signal.

\subsection{Region-based Analysis}
\label{sec:region}
For the region-based approach, the selected ROI is divided into several (five in our examples for the facial videos) sub-regions to preserve the signal intensity of global analysis but, on the other hand, analyze region-based differences. 
The regions are selected in a way that, despite object (or camera) position, at least one region can be captured by the camera to counteract possible occlusions.
For each region, the rPPG signal is extracted individually to determine the subject's HR $f_\mathit{HR}$ \cite{kossack2021} as well as the region-specific parameters presented in \cref{sec:parameter}. 

\subsection{Local Analysis}
\label{sec:local}
The local analysis resolves the rPPG signal on a pixel level. It allows identifying poorly perfused areas and the accurate differentiation of living tissue from non-living materials. Each spatial position within the image is analyzed. Beforehand, a reference region has to be defined to extract the reference rPPG signal and a global $f_\mathit{HR}$. If no reference region is present, an external HR measurement can also be used to receive $f_\mathit{HR}$.
As the influence of noise is stronger for local analysis, we calculate a Gaussian pyramid for each registered image. We chose a pyramid level consisting approx.~\SI{10000}{px} (i.e., \SI{100}{px} x \SI{100}{px}) and conducted an rPPG signal extraction as well as parameter calculation (see \cref{sec:parameter}) for each spatial position. The locally resolved parameters can then be used to visualize local perfusion behaviors, \eg, blood flow (see \cref{fig:localCorrFlap4Frames}).


\section{Datasets}
\label{sec:data}
We recorded two datasets, \textit{SurgTissue} and \textit{FaceMask}, for the evaluation of our perfusion assessment approach.

Dataset~I \textit{SurgTissue} consists of the operation situs of 14 oncological patients, who underwent tumor resection and defect reconstruction with a free fasciocutaneous flap for head-neck tumors (see \cref{fig:localCorrFlap4Frames}) as well as 27 patients receiving a kidney transplantation (see \cref{fig:localCorrKidney4Frames}). After reconstruction of the defect with the flap, respectively transplantation of the kidney, the artery as well as the accompanying veins were anastomosed (surgically connected). The reperfusion after opening the clamped artery was documented with an all-digital surgical imaging device with resolution of \SI{1920}{px} x \SI{1080}{px}, either a surgical microscope (ARRIscope, Munich Surgical Imaging, Germany) with \SI{60}{fps} or an endoscope (Karl Storz SE, Germany) with \SI{25}{fps}. The patient's vital signs were recorded continuously and synchronized with each video recording for validation reasons.

Dataset~II \textit{FaceMask} is a collection of RGB video recordings of masked and unmasked subjects. In total, 44 participants aged between 22 to 62 years, \SI{43}{\percent} female and \SI{57}{\percent} male, coming from diverse ethnic backgrounds (Hispanic, Asian, African, European) were captured using two different RGB cameras (35 subjects with ace acA2440-75uc, Basler AG, Ahrensburg, Germany and 9 subjects with RealSense D435, Intel Corp, CA, USA). Each person was recorded ten times for \SIrange{20}{30}{\second} each with resolution of \SI{2448}{px} × \SI{2048}{px} (ace with frame rate 25, 30, 40, or \SI{60}{fps}) or \SI{1280}{px} × \SI{720}{px} (RealSense D435 with \SI{30}{fps}). The subjects were illuminated with white light using LED panels (Diva-Lite L20X and L30X, Kino Flo, Inc., CA, USA) and varying light temperatures (\SI{2500}{K}, \SI{5000}{K}, \SI{9900}{K}).
Each participant was connected to a vital sign monitor (VitaGuard 3100, GETEMED, Germany) to measure ECG (via body electrodes, $f_s=\SI{256}{\hertz}$) and PPG (via finger clip sensor, $f_s=\SI{64}{\hertz}$) simultaneously to the video recording. 
The respective subject looked either straight into the camera or as well as moved the head slightly during the recording. First, we recorded the unmasked subject with and without movement. Subsequently, the subject was masked by applying full or partial facial coverings or thick layers of make-up. Finally, we recorded the masked participants again with and without movement. The partial coverage was applied to either the cheek, forehead, chin, or nose region using different variations (skin tones and sizes) of professional make-up equipment. Example images of the \textit{FaceMask} dataset are shown in the first column of \cref{fig:localMaps}. 


\section{Results and Applications}
\label{sec:app}
In the following, we present results for different applications. First, we show how the scale analysis can measure perfusion and blood flow during surgery. In the second step, we present region-based and local presentation attack detection (PAD).

\subsection{Intraoperative Perfusion Assessment}
In reconstructive and transplant surgery, flap respectively organ monitoring is crucial for early detection of perfusion problems. The survival of transplanted tissues anastomosed to suitable donor vessels depends on adequate tissue perfusion. Transplantation failure may result due to arterial or venous occlusion due to vasospasm, thrombosis, external compression, vessel kinking, or hematoma formation \cite{Chae2015}. Furthermore, timely re-exploration can significantly increase the rate of compromised tissue salvage. Consequently, detecting early signs of deterioration and the necessary correction is only possible with close monitoring~\cite{Chang2013SalvageRO,Yang2014}. Although several objective monitoring techniques have been reported in the past~\cite{Swartz1988,Hashimoto2007,Nagata2014} and a growing need for a reliable and examiner-independent assessment is depicted, the current gold standard for perfusion monitoring is still based on subjective clinical assessment~\cite{Chae2015}. As a result, since not all parts of the vessel remain visible and slight discoloration is invisible to the human eye, suboptimal positioning as well as slight kinks in the course of the vessel remain undetected though impairing organ function. Analyzing the \textit{SurgTissue} dataset shows statistically distinct behavior for perfused and non-perfused tissue parts, allowing to specify relevant time points and tissue behaviors.

\begin{figure}[ht]
	\centering
	\includegraphics[width=1.0\linewidth]{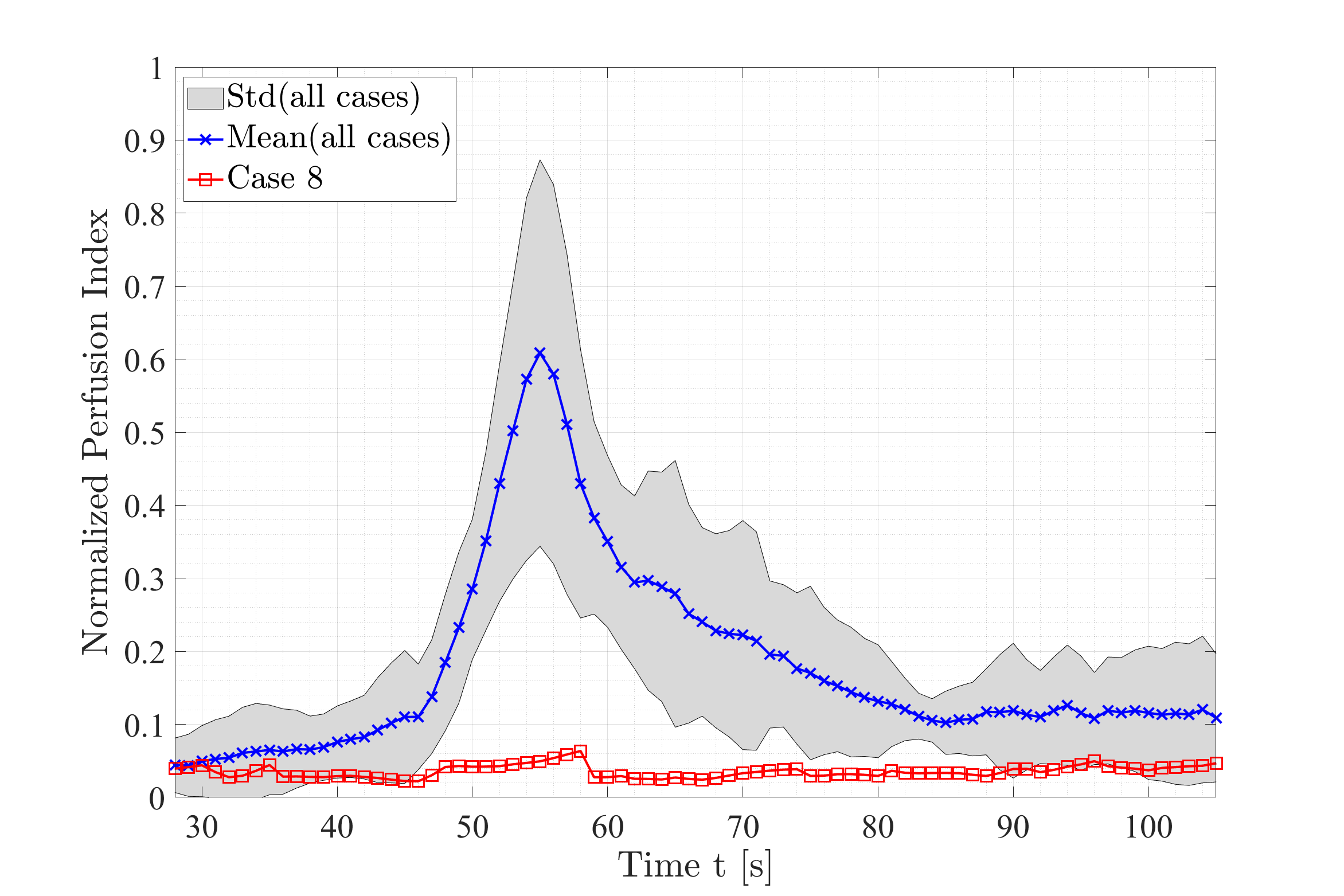}
	\caption{Normalized perfusion index showing the exact time of reperfusion of transplanted flaps. Maximum and half-width of the perfusion peak depend on the individual temporal opening of the external carotid artery. No reperfusion is detected for case $\#8$ (red curve) suggesting insufficient blood flow.}
	\label{fig:piGlobAll}
\end{figure}

\begin{figure}[ht]
	\centering
	\includegraphics[width=1.0\linewidth]{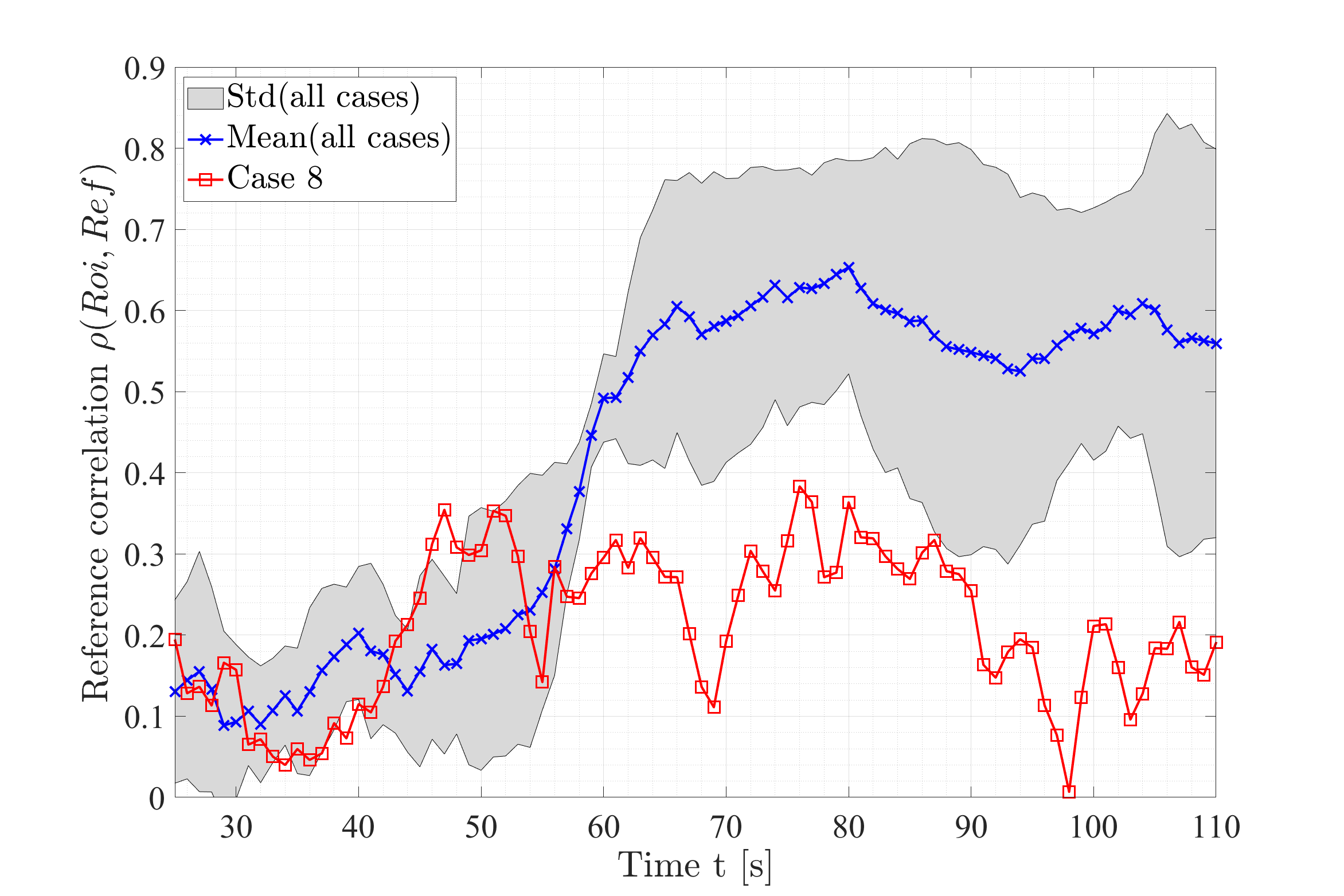}
	\caption{Correlation of the pulse-induced rPPG signal for transplanted flaps around the time of reperfusion ($\sim$\SI{50}{\second}) with a defined reference region. The higher the correlation value, the closer the perfusion in the flap matches the healthy perfused reference region. For case \#8 (red curve), no perfusion is detectable in the flap resulting in a permanently low correlation value.}
	\label{fig:corrGlobAll}
\end{figure}

\subsubsection{Global and Region-based Assessment}
For all free fasciocutaneous flap patients in the \textit{SurgTissue} dataset, it is possible to quantify how accurately the HR can be determined from the video data using the recorded ground truth comparison values. Further, for all patients, where the flap transplantation went successfully (13 out of 14 cases), it is possible to detect region-based perfusion changes in the transplanted tissues.
From the extracted rPPG signal of the flap region, the $\mathit{PI}$ shows a significant change of perfusion, i.e., the reperfusion of the tissue can be detected, see \cref{fig:piGlobAll}. This event of reperfusion produces a peak, reaching its maximum $t_\mathit{PI} = \SI{15}{second}$ after the documented reperfusion event. The reference correlation $\rho$ between the flap and a reference skin region increases in parallel to the increasing $\mathit{PI}$, reaching its maximum at $t(\mathit{PI}_{max})$, see \cref{fig:corrGlobAll}.

For the case \#8, no characteristic parameters could be derived intraoperatively, cf.~red curve in \cref{fig:piGlobAll} and \cref{fig:corrGlobAll}. This flap showed no vitality after inspection on the 3rd postoperative day and the necrotic flap tissue was removed 5 weeks postoperatively.

\subsubsection{Local Assessment}
A good visualization of blood flow propagation and representation of the signal quality of each spatial position would allow the surgeon to make a robust quantitative diagnosis. In addition, it could reveal transplantation failures at an early stage without the need for additional complex and invasive methods.

\begin{figure}[htb]
	\centering
	\includegraphics[width=1.0\linewidth]{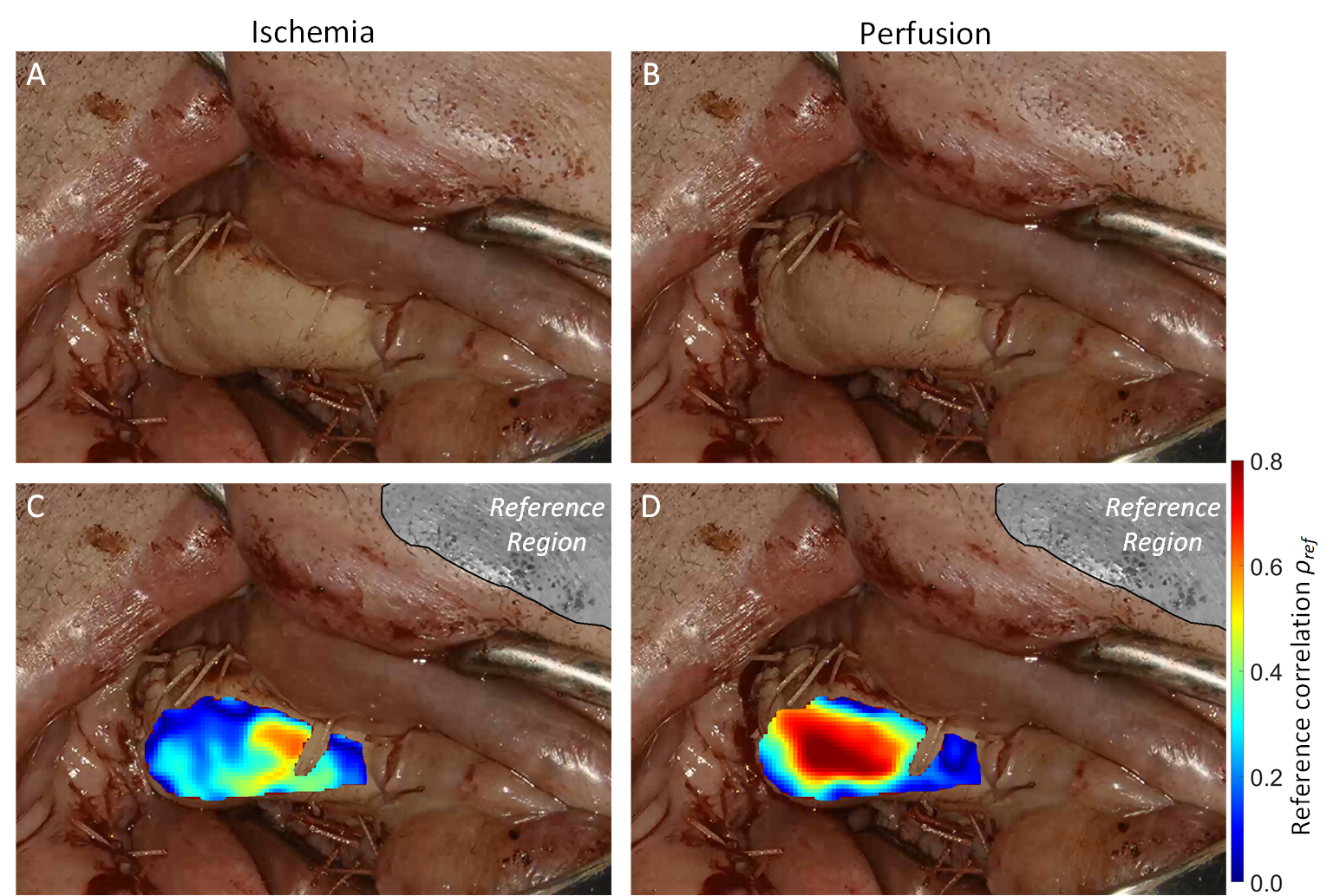}
	\caption{Local reference correlation between the pixel-wise position within the flap and a reference signal of a well-perfused skin area (highlighted in gray) at two points in time; before reperfusion (A) and after reperfusion (B). The improvement of the reference correlation of the flap before reperfusion (C) to a time point after (D) can be visualized with the local perfusion assessment.}
	\label{fig:localCorrFlap4Frames}
\end{figure}

\begin{figure}[htb]
	\centering
	\includegraphics[width=1.0\linewidth]{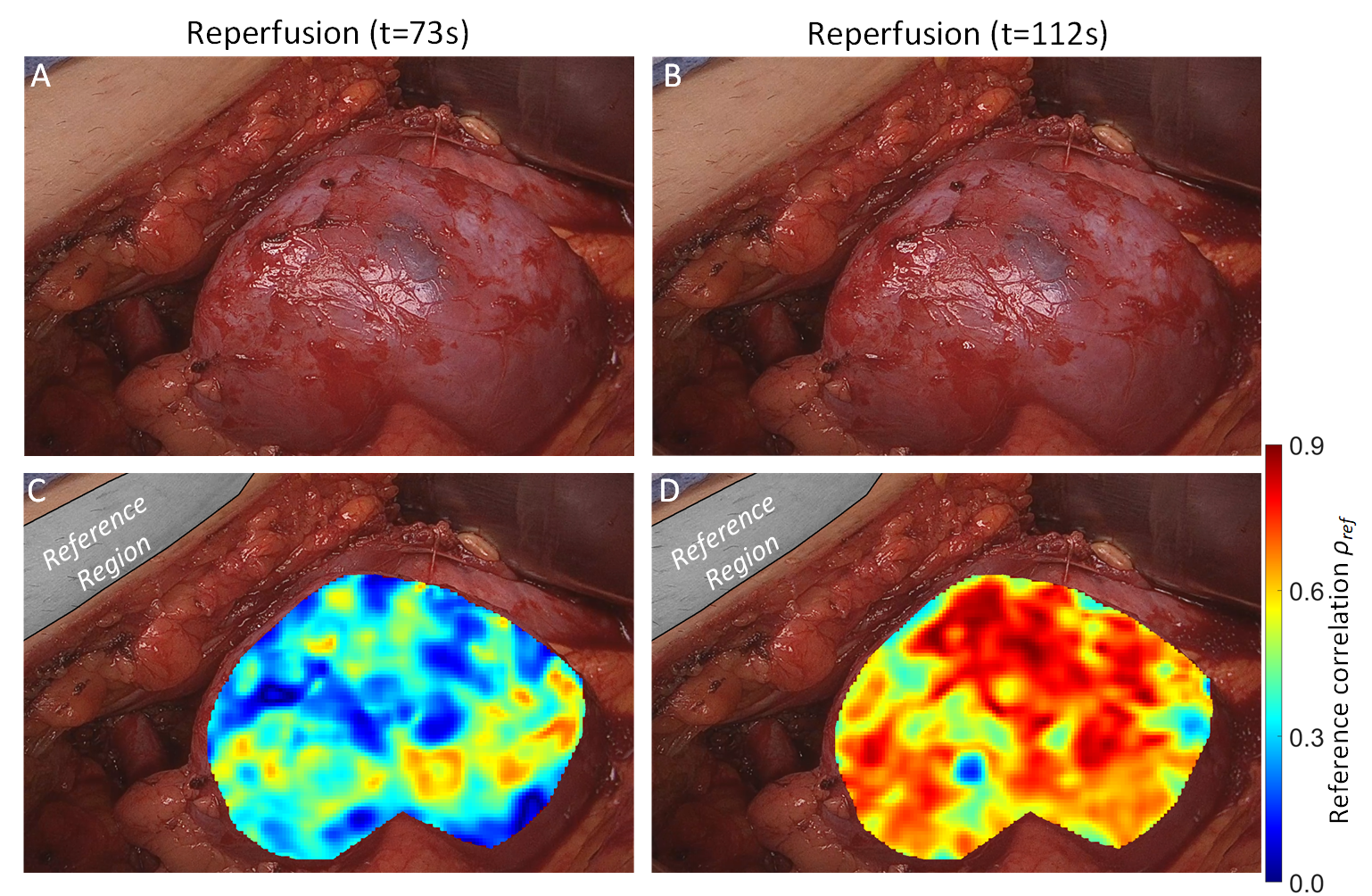}
	\caption{This figure shows the reperfusion during the implantation of a donated kidney. In the input images of the kidney, no perceptible changes can be detected after \SI{73}{\second} (A) and \SI{112}{\second} (B) after reperfusion. However, by the visualization using local analysis and corresponding reference correlation map (bottom row), an evident change and improvement of perfusion between (C) and (D) can be observed. The reference skin region is marked in gray.
}
	\label{fig:localCorrKidney4Frames}
\end{figure}

An augmented overlay of the different parameters, which have been determined locally, allows to visually highlight relevant tissue behaviors during critical processes. The normal RGB view on a transplanted region shortly after reperfusion shows no difference compared to the non-perfused appearance, cf.~\cref{fig:localCorrFlap4Frames} and \cref{fig:localCorrKidney4Frames} (top row). However, the local reference correlation shows a clear improvement in perfusion over time with specific local variations.

\subsection{Presentation Attack Detection via Perfusion Analysis}
The goal of a presentation attack is to impersonate a person's identity through the presentation of photos or wearing a mask such that an identification system allows the attacker to pass through the system with a false identity. The \textit{FaceMask} dataset includes attacks with full face masking, cf.~\cref{sec:data}. Such full-face masks with the authorized person's biometric feature have the disadvantage of being sometimes easily recognized as the masks do not move like a real face, and the reflection of light might differ from human skin. More advanced attacks involve partial masks altering a person's face so that the attacker's biometric features match the target victim's. These masks can be blended into the face's overall appearance with make-up equipment to create a natural look. In the dataset, such partial mask attacks are simulated using handmade partial masks made of gelatin, latex, latex milk, wax, and Artex silicone. An application of perfusion assessment to detect masks (i.e., presentation attacks) in video sequences is presented in the following. 

\subsubsection{Region-based PAD}

	\begin{figure*}
		\centering
			\includegraphics[width=0.85\linewidth]{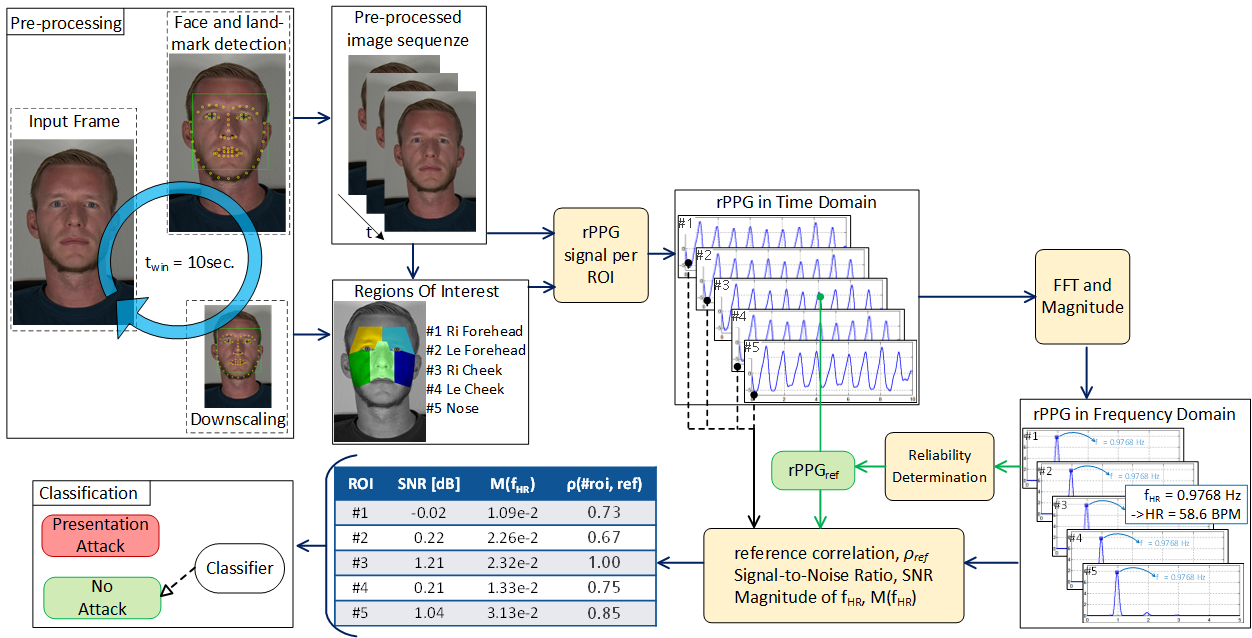}
		\caption{Illustration of our presentation attack detection approach based on region-based perfusion assessment. After pre-processing (face and landmark detection and downscaling), the region-based rPPG calculation is applied to the pre-processed image sequence. The resulting signals are transformed into the frequency domain, where the heart rate is determined. Subsequently, the reliability determination \cite{kossack2021} is performed, and thus the reference signal $\mathit{rPPG}_{\mathit{ref}}$ is determined. The parameters SNR, $M(f_\mathit{HR})$, and $\rho_{ref}$ can be calculated for each region. Based on the calculated parameters, the SVM classifier classifies between presentation attack and no attack.}
		\label{fig:regionPAD}
	\end{figure*}   

\cref{fig:regionPAD} shows the major steps of the proposed region-based PAD. To evaluate this approach, recordings of the \textit{FaceMask} dataset have been used, containing simulated attacks with full face masks out of latex and foam latex. Randomly, 43 video recordings are selected; 17 of them are showing presentation attacks. The length of each video is approx.~\SI{30}{\second}. The position of the face is determined using face detection based on the dlib algorithm \cite{king2009} extracting 68~landmarks per face. Based on these landmarks, the face is segmented into five symmetric regions covering the major parts of the face, excluding the eyes since they mostly do not show visible skin. \cref{fig:regionPAD} shows the regions: right forehead, left forehead, right cheek, left cheek, and nose. The mouth and chin areas are not analyzed due to low signal quality \cite{Kwon2015}. The pre-processing steps for facial videos described in \cref{sec:region}, yield \SI{10}{\second} long image sequence and the corresponding facial landmarks for each frame which determine the facial ROIs. To determine the HR of the recorded subject, the region with the most robust signal based on a reliability determination has been selected~\cite{kossack2021} as reference signal $\mathit{rPPG}_{\mathit{ref}}$.

For each region and sliding window, the three parameters ($\mathit{SNR}$, $M(f_\mathit{HR})$, $\rho_{\mathit{ref}}$) are calculated, cf.~\cref{fig:regionPAD}, resulting in 2216 ground truth samples. A cubic SVM classifier is trained to detect presentation attacks. The data are split into training and test set (training: \SI{80.3}{\percent} (1780 samples), testing: \SI{19.7}{\percent} (436 samples)). Thereby, the data set was randomly split with the constraint that no subject of the test set appears in the training set and vice versa.
We use cross-validation to protect the classifier against over-fitting to the training data (10 folds). The classifier is tested against the test data and reaches an accuracy of \SI{96.8}{\percent}. \cref{tab:confusion} shows the confusion matrix visualizing the performance of the classifier.

\begin{table}[!ht]
    \centering
    \caption{Confusion matrix of full-mask presentation attack classification (no attack = 0, presentation attack = 1).}
\noindent
\setlength\tabcolsep{0pt}
\begin{tabular}{rcccr}
 &   & \multicolumn{2}{c}{Predicted Class} & \\
 &   & \textbf{0} & \textbf{1} & \textbf{total} \\
\multirow{2}{*}{\parbox[c]{0.5cm}{\rotatebox{90}{\textbf{Actual Class}}}}
 & 0 & \fbox{\parbox[c][0.9cm]{0.9cm}{\centering 200}} & \fbox{\parbox[c][0.9cm]{0.9cm}{\centering 4}} & 204 \\
 & 1 & \fbox{\parbox[c][0.9cm]{0.9cm}{\centering 11}} & \fbox{\parbox[c][0.9cm]{0.9cm}{\centering 221}} & 232 \\
 & total & 211 & 225 &
\end{tabular}
\label{tab:confusion}
\end{table}

\subsubsection{Local PAD Visualization}
\begin{figure}[htb]
	\centering
	\includegraphics[width=1.0\linewidth]{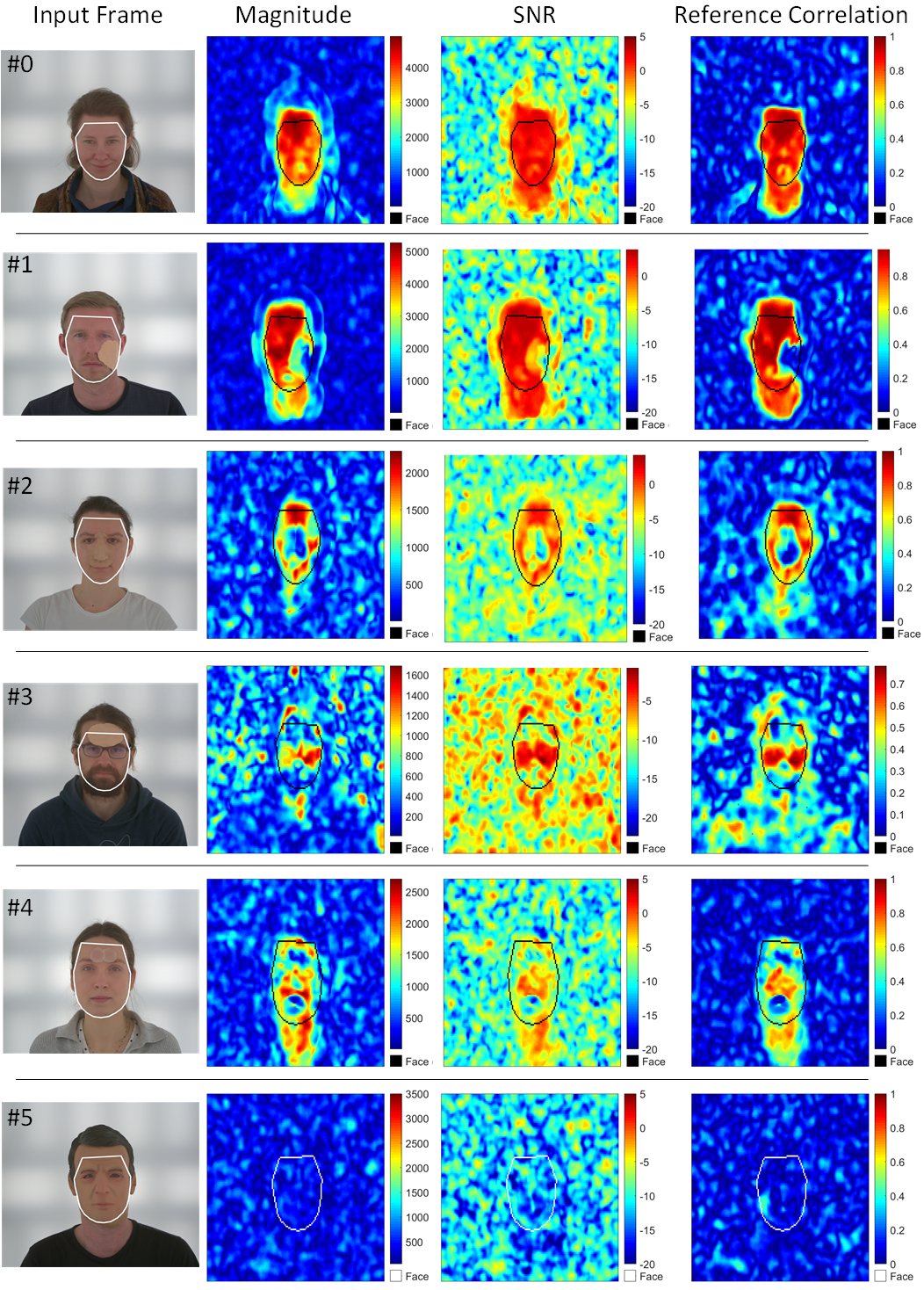}
	\caption{This figure shows the resulting maps of the local analysis of different subjects wearing \#0 no mask, \#1 cheek mask, \#2 nose mask, \#3 forehead mask, \#4 forehead mask, and \#5 full face mask. In the respective maps, the pixel positions in red are assigned to a high value and blue to a low value for the respective parameter. The contour drawn in each plot shows the subject's face position.}
	\label{fig:localMaps}
\end{figure}

Local perfusion assessment can be used to visualize more details of the local perfusion within the human face. We use the local analysis described in \cref{sec:local} and visualize our approach for local perfusion assessment. This visualization can be used for PAD assistance to detect and localize partial masks. The \textit{FaceMask} dataset (see in \cref{sec:data}) simulates presentation attacks with partial masks. 
These masks cover single or multiple facial regions like the nose, cheeks, or forehead. \cref{fig:localMaps} shows four examples of different face coverings and one non-covered face. It shows the input frame, the magnitude-, SNR- and reference correlation-maps. All five attacks are clearly visible within each corresponding map. With the local rPPG perfusion assessment, we can determine if a person is wearing a mask and localize the position of this mask coverage.

\section{Discussion}
\label{sec:discussion}
The rPPG is a non-contact vital parameter determination method and has been used for objective intraoperative perfusion assessment of transplanted tissues. The overall goal of the considered surgical procedures is to achieve a similar blood flow in the transplanted tissues compared to the reference region. Sufficient reperfusion in transplanted tissue resulting from opening the artery is clearly detectable by the PI peak and the increase in correlation of the pulse-induced rPPG signal. Further, we could show that rPPG is able to detect critically perfused tissues before clinical ischemia occurs.

The correlation quantifies the similarity of both rPPG signals, which allows clear conclusions about the perfusion of the transplanted tissue. Therefore, an objective and reproducible perfusion assessment can be made region-based as well as locally based on the introduced parameters. This gives to surgeon additional relevant information about physiological parameters and perfusion problems, which would manifest clinically much later so that a possible intervention would be already too late. For this reason, such a non-invasive continuous monitoring approach using existing imaging techniques such as surgical microscopes and endoscopes would be an excellent improvement for various medical applications, such as intraoperative tissue perfusion measurements as well as postoperative wound assessment. 
All results calculated for the \textit{SurgTissue} dataset and presented in this paper have been reviewed and verified by several medical experts (for free fasciocutaneous flap and kidney transplantation). 

As shown in \cref{fig:localMaps}, the partial and full-face masks can be located in a person's face by inspecting the magnitude, SNR, and reference correlation maps generated via the local analysis. However, the figure also shows for subject \#3 that his dense beard appears as an occlusion, which in fact is an occlusion of skin. Thus, facial hair makes it difficult robustly detect presentation attacks, as it has to be distinguished between normal occlusions such as beard and glasses and attacks with partial masks. Because of this challenge, as well as it has been reported that the chin generally shows less signal strength \cite{Kwon2015}, we decided to ignore the chin region for the region-based PAD. With the region-based approach, we were able to detect full-face coverage. Compared to the local analysis, the advantage of the region-based method is reducing processing time, making real-time application possible.

\section{Conclusion}
\label{sec:conclusion}
We presented a method for rPPG-based perfusion assessment in visible human tissue from RGB video-based. We use a plane-orthogonal-to-skin remote photoplethysmography technique in order to extract the rPPG signal and derive perfusion-relevant parameters at three different scales, offering flexibility for different applications. 

We presented two example applications for intraoperative tissue perfusion monitoring in patients undergoing transplantation or reconstruction surgery. Our approach yields objective and reproducible results, verified by medical experts. We will perform further clinical studies to evaluate our approach for continuous intraoperative as well as postoperative tissue monitoring. Existing digital imaging units, such as a surgical microscope or endoscope could be easily expanded \cite{Wisotzky2019} to be used for tissue monitoring. 

In addition to applications in the medical field, we presented an application to presentation attack detection to authentication systems. In our experiments, attacks with partial- and full-face masks could robustly been detected.


\section{Acknowledgment}
\label{sec:ack}
The work in this paper has been funded in part by the H2020 projects \textit{D4Fly} under grant number 952147, by the German Federal Ministry of Education and Research (BMBF) projects \textit{KIPos} under grant number 16SV8602, and \textit{MultiARC} under grant number 16SV8061.

{\small
\bibliographystyle{ieee_fullname}
\bibliography{egbib}
}

\end{document}